\documentclass[runningheads]{llncs}

\usepackage{accv}

\usepackage{accvabbrv}

\usepackage{graphicx}
\usepackage{booktabs}

\usepackage[accsupp]{axessibility}  

\usepackage{xcolor,colortbl}

\usepackage[pagebackref,breaklinks,colorlinks,citecolor=accvblue]{hyperref}

\usepackage{orcidlink}

\begin{document}

\title{HAHA: Highly Articulated Gaussian Human Avatars with Textured Mesh Prior} 

\titlerunning{HAHA}

\author{David Svitov\inst{1, 2}\orcidlink{0009-0009-9116-0416} \and
Pietro Morerio\inst{2}\orcidlink{0000-0001-5259-1496} \and
Lourdes Agapito\inst{3}\orcidlink{0000-0002-6947-1092} \and \\
Alessio {Del Bue}\inst{2}\orcidlink{0000-0002-2262-4872}}

\authorrunning{D.~Svitov et al.}

\institute{
Università degli Studi di Genova, Italy \and
Istituto Italiano di Tecnologia (IIT) Genoa, Italy \\
\email{\{david.svitov, pietro.morerio, alessio.delbue\}@iit.it} \and
Department of Computer Science, University College London\\
\email{l.agapito@cs.ucl.ac.uk}}

\maketitle

\begin{figure}[tb]
  \centering
  \includegraphics[width=0.45\paperwidth]{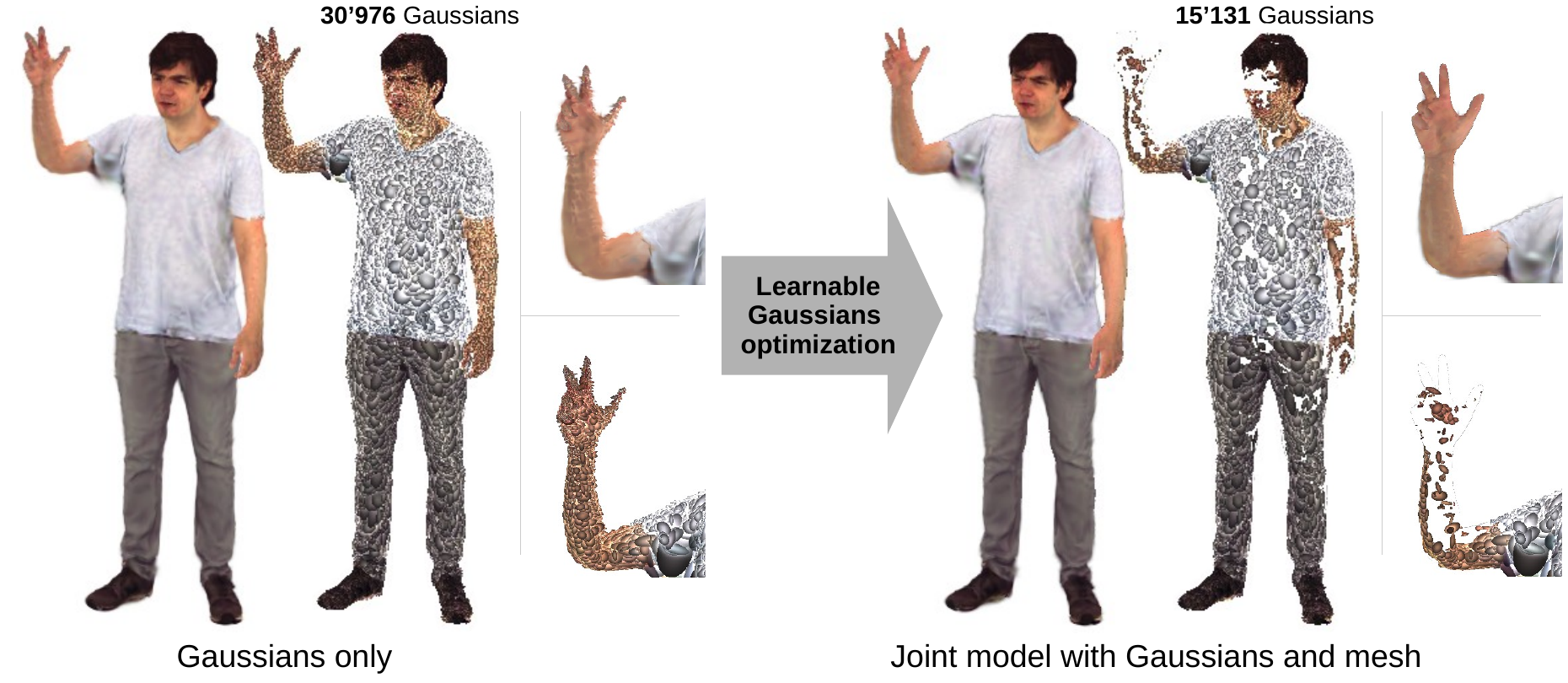}
  \caption{\textbf{Optimizing the number of Gaussians.} HAHA jointly optimizes a Gaussian splatting model with a textured mesh to improve the photometric quality of the avatars. The method filters out superfluous Gaussians in a learnable, unsupervised manner. As a result, we can more efficiently and better animate highly articulated parts of a body.}
  \label{fig:teaser}
\end{figure}

\begin{abstract}
  We present HAHA - a novel approach for animatable human avatar generation from monocular input videos. The proposed method relies on learning the trade-off between the use of Gaussian splatting and a textured mesh for efficient and high fidelity rendering. We demonstrate its efficiency to animate and render full-body human avatars controlled via the SMPL-X parametric model. Our model learns to apply Gaussian splatting only in areas of the SMPL-X mesh where it is necessary, like hair and out-of-mesh clothing. This results in a minimal number of Gaussians being used to represent the full avatar and reduced rendering artifacts. This allows us to handle the animation of small body parts, such as fingers, that are traditionally disregarded. We demonstrate the effectiveness of our approach on two open datasets: SnapshotPeople and X-Humans. Our method demonstrates on par reconstruction quality to the state-of-the-art on SnapshotPeople, while using less than a third of Gaussians. HAHA outperforms previous state-of-the-art on novel poses from X-Humans both quantitatively and qualitatively.
\end{abstract}

\section{Introduction}
\label{sec:intro}
The task of creating photo-realistic animated objects has always been of paramount importance in 3D computer vision. High-fidelity animated objects are widely used in real-time applications, ranging from computer games to online telepresence systems \cite{jones2021belonging, lexfridman}. In recent years the interest in the field has increased due to the emergence of devices for virtual \cite{meta} and augmented \cite{apple} reality. Traditionally, the central aspect of the task is the creation of a human avatar as it has a wide range of uses and digital replicas are essential for online human-to-human interaction. Therefore, our work concentrates on rendering animated photo-realistic human avatars.

To date, several options are available to generate human avatars for what concerns input data. To get the best quality many methods rely on multi-view data \cite{bashirov2024morf, chen2021animatable, zhao2022high, jiang2022instantavatar, grigorev2021stylepeople}. However, complex acquisition systems such as a multi-camera capture setup \cite{isik2023humanrf} or a 3D scanner \cite{Texel} are required to collect such data. On the other hand, some methods \cite{svitov2023dinar, he2021arch++, saito2019pifu, alldieck2022photorealistic} use a single image of a person as input, which overcomplicates the task with the necessity to restore unobserved regions of the body. Eventually, the most convenient way is to generate avatars from \textbf{monocular videos}. Using a monocular video provides a trade-off between the complexity of obtaining input data and the quality of the avatar.

In the last few years, monocular video avatars have been represented using explicit \cite{alldieck2019learning, alldieck2018detailed, alldieck2018video} or implicit \cite{chen2021animatable, peng2021neural, jiang2023instantavatar, yu2023monohuman} geometry. Recently, a novel method for representing 3D objects has appeared - \textbf{Gaussian splatting} (3DGS) \cite{kerbl20233d} - where the scene is represented as a set of parametrized Gaussians, which are projected onto the screen surface during rendering. The most recent methods for human avatars \cite{lei2023gart, zielonka2023drivable, hu2023gaussianavatar, qian2023gaussianavatars, jena2023splatarmor, qian20233dgs, chen2023monogaussianavatar, dhamo2023headgas, wang2023gaussianhead, hu2023gauhuman} indeed utilize 3DGS for rendering. These works cover an extensive range of tasks, from head avatars to multi-view full-body avatars. With this representation, temporal consistency is improved over implicit methods, and out-of-mesh details are more accurately conveyed than with traditional explicit methods. To drive the animation, previous works traditionally employ parametric models \cite{loper2023smpl, FLAME:SiggraphAsia2017, pavlakos2019expressive} of the human body. This way, they can control the shape and pose of the body via learnable parameters.

A common drawback of existing Gaussian-based methods is that they require \textbf{a large number of Gaussians} to represent a human avatar. Especially if we need to animate high-frequency details such as fingers. These regions of the body could require a tremendous amount of Gaussians to look realistic enough. Up to 200'000 for previous approaches \cite{hu2023gaussianavatar} to represent an avatar. This in turn leads to an increase in the required memory. Moreover, if we need to improve the details of the resulting avatar, we can only increase the number of Gaussians. This could be a bottleneck when we want to render a scene with many avatars (\eg for a game or a movie). Another issue with monocular video-based Gaussian avatars is that video frames from a single camera are often insufficient to generalize to novel views and poses efficiently. Mesh-based explicit methods \cite{alldieck2019learning, alldieck2018detailed, alldieck2018video} circumvent this issue by strongly relying on mesh geometry, whereas Gaussians tend to overfit. However, these methods struggle to reconstruct loose clothes and hair accurately.

In this work, we introduce \textit{Highly Articulated Gaussian Human Avatars with Textured Mesh Prior (HAHA)}. While existing approaches focus on using the mesh-based approach \cite{Zheng2023AvatarReXRE} or Gaussian-based approach \cite{moreau2024human}, we target to take the best from both representations. Our main idea is to \textbf{learn to use the appropriate number of Gaussians relying on a textured mesh where possible} (Fig. \ref{fig:teaser}). We attach Gaussians to the mesh surface only at the points where it is necessary to represent out-of-mesh details. For the mesh, we use SMPL-X \cite{pavlakos2019expressive} parametric human model with articulated fingers and face, and in contrast with previous approaches that use SMPL \cite{loper2023smpl} we aim to control fingers animation as well as the bigger joints. Areas not covered with the Gaussians are represented as a textured mesh surface that is more efficient to store. Using such a mesh, we significantly reduce the number of Gaussians in the areas of the hands and face (Fig. \ref{fig:teaser}). Overall, we reduce the amount of Gaussians \textbf{up to three times} for the whole avatar, resulting in $\times2.3$ reduced storage costs.

We obtain an avatar with a three-stage pipeline. During the first two stages, we learn Gaussian and textured mesh representation of the avatar. In the final stage, we estimate which Gaussians to remove in an unsupervised manner. We proposed the mechanism for the combined differentiable rendering of Gaussians and a mesh, which allows us to adjust Gaussians' parameters 
based on the final rendering of the avatar. 

We propose several regularization techniques to encourage \textit{HAHA} to remove as many Gaussians as possible without affecting the quality of the avatar. Following 3DGS \cite{kerbl20233d} our Gaussians have trainable opacity and we delete them when it is lower than a threshold. We use two regularizations balancing each other to control Gaussians' opacity during training. While the first pushes opacity down, the second controls out-of-mesh detail preservation. This way, we find a learnable trade-off in using Gaussians and a textured mesh. To train \textit{HAHA} in such a manner, we only need input video frames with the provided SMPL-X fits without any additional labels.

In our experiments, we show that \textit{HAHA} reaches quantitative metrics on par with state-of-the-art methods \cite{lei2023gart, qian20233dgs, hu2023gaussianavatar} on the open SnapshotPeople dataset \cite{alldieck2018video}, while better generalizing to novel poses and views. Using videos from the X-Humans dataset \cite{shen2023xavatar}, we demonstrated that \textit{HAHA} allows us to animate fingers with higher quality than state-of-the-art. We demonstrate that our method, both qualitatively and quantitatively, outperforms state-of-the-art methods on agile X-Humans data, while at the same time, it allows us to reduce the number of Gaussians.

The main contributions of the work are the following:

\begin{itemize}
 \item We first propose the use of Gaussians in combination with textured mesh to increase the efficiency of rendering human avatars;
 \item We develop an unsupervised method for significantly reducing the amount of Gaussians in the scene through the use of textured mesh;
 \item We demonstrate that our method can efficiently handle the animation of hands and other highly articulated parts without the need for any additional engineering.
\end{itemize}

\section{Related Work}
\label{sec:related}

\textbf{Human parametric models.} Parametric models such as SMPL \cite{loper2023smpl} or FLAME \cite{FLAME:SiggraphAsia2017} are widely used in human avatars \cite{alldieck2019learning, alldieck2018detailed, jiang2023instantavatar, chen2021animatable, lei2023gart, qian20233dgs, grassal2022neural, duan2023bakedavatar} to control pose and shape. The parametric model gets as input vectors of the pose and shape parameters and produces the mesh. Such a mesh is posed using linear blend skinning (LBS) when the pose vector controls pose-dependent body transformation. The resulting mesh may be used to transform an avatar to the canonical pose \cite{jiang2023instantavatar, chen2021animatable} or to directly form an avatar appearance \cite{alldieck2019learning, alldieck2018detailed}.

Researchers traditionally use SMPL to get avatars, while the most flexible parametric model is SMPL-X \cite{pavlakos2019expressive}. This model allows one to additionally control finger joints and facial expressions. Therefore, it is more useful for practical use cases of avatars. \textit{HAHA} uses as input SMPL-X's parameters corresponding to video frames. One can get SMPL-X's pose and shape parameters from input images using SMPLify-X\cite{pavlakos2019expressive} method or one of the recent feed-forward methods \cite{kocabas2020vibe, sun2021monocular}.

\textbf{Gaussian splatting avatars.} 3DGS \cite{kerbl20233d} appeared recently as a novel method for explicit scene representation. The method represents a scene as a collection of 3D Gaussians and their associated photometric information. These Gaussian splats on the camera image surface produce a rendered image during rendering. 3DGS demonstrated its efficiency for static scene representation \cite{jiang2023gaussianshader, Huang2024OptimalPF, Yu2023MipSplatting, lee2024deblurring} as well as for dynamic scenes \cite{luiten2023dynamic, kratimenos2023dynmf, jiang2023hifi4g, Duan20244DGS}. Recent methods \cite{lei2023gart, zielonka2023drivable, hu2023gaussianavatar, qian2023gaussianavatars, jena2023splatarmor, qian20233dgs, chen2023monogaussianavatar, dhamo2023headgas, wang2023gaussianhead, hu2023gauhuman} use 3DGS for rendering photo-realistic human avatar in different operational scenarios. They generate avatars based on the multi-view data \cite{li2023animatable, zheng2023gps, pang2023ash} or a monocular-video \cite{lei2023gart, jena2023splatarmor, hu2023gaussianavatar, qian20233dgs} input. Using 3DGS for avatar rendering allows authors to obtain temporally consistent animated rendering with better metrics value. 

Current state-of-the-art methods use SMPL to drive animation in Gaussian-based human body rendering. For instance, GART \cite{lei2023gart} represents Gaussians in the canonical space and uses skeletons with learnable LBS weights to animate them.  To handle out-of-mesh details, they proposed to create additional bones. 3DGS-Avatar \cite{qian20233dgs} sets Gaussians in the canonical space and models non-rigid deformations with a learnable MLP network. The authors also applied as-isometric-as-possible regularization \cite{kilian2007geometric} to the Gaussians to preserve geometric consistency. GaussianAvatar \cite{hu2023gaussianavatar} enforces inductive bias by using CNN to generate Pose Features in SMPL's texture space. GaussianAvatar optimizes this model and the SMPL pose to compensate for SMPL's inaccuracy. SplatArmor \cite{jena2023splatarmor} embeds Gaussians to the SMPL surface in the canonical space. They use Neural Color Field to preserve inductive bias and use MLP to predict non-rigid transformation.

Another approach \cite{Waczynska2024GaMeSMA, qian2023gaussianavatars, saito2023relightable, xiang2023flashavatar, pang2023ash} is to attach Gaussians to the mesh's polygons. 
But so far, such methods focus on mostly rigid objects (\eg heads) or use multi-view data as input. This work demonstrates the efficiency of such an approach for the monocular video-based full-body human avatars.

Several previous works \cite{Zheng2023AvatarReXRE, shen2023xavatar} solve the task of generating human avatars with articulated finders using multi-view data.  AvatarReX \cite{Zheng2023AvatarReXRE} uses a separate parametric model to process hands. As input, they accept a multi-view video of a person. X-Avatar \cite{shen2023xavatar} uses a part-specific deformer network to handle the hands. As input X-Avatar gets 3D scans or RGB-D video with depth information. Both methods reconstruct an avatar as a textured mesh that in general leads to more blurred results than 3DGS. In contrast to the previous works, we use only monocular RGB data.

\textbf{Texture-based avatars.} The classical approaches generate video-based human avatars using textured meshes \cite{alldieck2019learning, alldieck2018detailed, alldieck2018video}. A mesh with RGB texture allows faster rendering with minimal artifacts, but the drawback of such an approach is the lack of out-of-mesh details. Existing methods try to circumvent this issue by predicting offsets to the SMPL mesh vertices. However, such an approach is limited by mesh topology as we can not represent enough details where the mesh grid is sparse.

To improve the textured mesh approach, researchers proposed the neural-texture rendering technique Deferred Neural Rendering (DNR) \cite{thies2019deferred}. In this approach, textures contain an arbitrary number of channels and can be interpreted as matrices of features. After rasterization, the method applies U-Net-like architecture to transform image channels to RGB. Several methods \cite{grigorev2021stylepeople, bashirov2024morf, raj2021anr, zhao2022high} build avatars using neural-textures. This allows them to represent more details than the RGB texture, especially out-of-mesh ones (\eg loose clothing). However, such methods are prone to temporal inconsistency and flickering during animation.

In this work, we research the new task of merging a novel Gaussian-based approach with a classical RGB texture-based. This allows us to reduce the number of Gaussians and, therefore, reduce memory requirements to store an avatar. Utilizing textured mesh where possible helps us reduce the number of artifacts connected with redundant Gaussians while remaining Gaussians represent out-of-the-mesh elements of avatars. Thus, we leverage the pros from both representations.

\section{Method}
\label{sec:method}

\begin{figure}[tb]
  \centering
  \includegraphics[width=0.55\paperwidth]{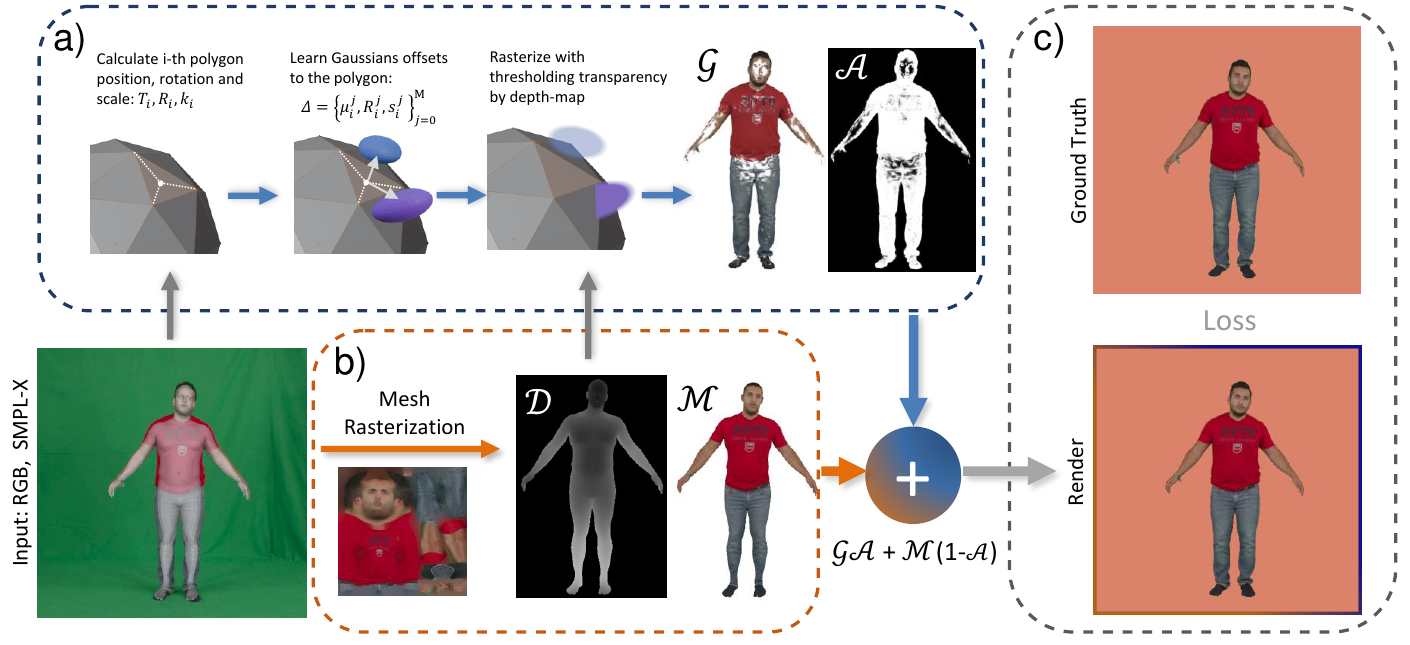}
  \caption{\textbf{Scheme of our approach.} a) We attach Gaussians to mesh polygons as described in Section \ref{sec:splatting} and rasterize them conditioned on depth map $\mathcal{D}$ into RGB image $\mathcal{G}$ and alpha map $\mathcal{A}$. 
  b) We train RGB texture for SMPL-X and rasterize mesh to RGB image $\mathcal{M}$ and depth map $\mathcal{D}$. c) During training and inference we merge rasterizations of Gaussians $\mathcal{G}$ and mesh $\mathcal{M}$, based on the trainable transparency map $\mathcal{A}$ of Gaussians.}
  \label{fig:scheme}
\end{figure}

Our pipeline comprises three stages. In the first (Fig. \ref{fig:scheme} (a)) stage, we learn a full-Gaussian representation of the avatar and fine-tune SMPL-X's poses and shapes for training frames. As a result, we get an avatar represented with Gaussians as in previous state-of-the-art approaches having a fixed initial set of Gaussians (\ie $N=20908$). In the second stage (Fig. \ref{fig:scheme} (b)), we use resulting SMPL-X meshes with the provided UV-map to learn RGB texture. Thus we obtain textured avatars without any out-of-mesh details but efficient to render and store. In the last stage, we merge these two avatars and learn to remove some Gaussians without losing quality (Fig. \ref{fig:scheme} (c)). To figure out which Gaussians to delete we perform combined rendering of the avatar and fine-tune Gaussians opacity. Further in this section, we describe these three stages in more detail.

\subsection{Gaussian Avatar Preliminaries}
\label{sec:splatting}
First, we describe how we set Gaussians on the SMPL-X mesh surface.
For each mesh's polygon, we calculate the coordinates of its center $T_i$, the quaternion rotation $R_i$, and the scale $k_i$ (Fig. \ref{fig:scheme} (a)). Then we calculate the parameters of the $N$ Gaussians $\Delta_i = \{\mu_i, r_i, s_i, c_i, o_i\}$ attached to each SMPL-X's polygon referred as $i$. Here $\mu_i, r_i, s_i$ are the Gaussian's translation, rotation, and scale offsets relative to $i$-th polygon parameters $\{T_i, R_i, k_i\}$, while $c_i$ and $o_i$ are the color and opacity properties, respectively. Similar to \cite{qian2023gaussianavatars} we perform a subdivision of Gaussians while maintaining the attachment to the parent polygon: $\Delta_i = \{\mu_i^j, r_i^j, s_i^j, c_i^j, o_i^j\}_{j=0}^{M_i}$ (Fig. \ref{fig:scheme} (a)). Thus, the final Gaussians pose and shape parameters are calculated as offsets to the corresponding $i$-th polygon parameters $\{T_i, R_i, k_i\}$ as follows:
\begin{equation}
r' = Rr  \hspace{20 mm}   \mu' = kR\mu + T   \hspace{20 mm}    s' = ks. 
\end{equation}

Several works \cite{hu2023gaussianavatar, li2023animatable, saito2023relightable, lei2023gart} demonstrated the effectiveness of using neighboring Gaussians information, similar to convolution inductive bias in the Convolutional Neural Networks. Such a technique increases the similarity between neighboring Gaussians and reduces the number of artifacts. Following \cite{lei2023gart}, we apply KNN regularization for Gaussians to constrain the transformation and appearance of neighbors. To further improve the avatars' quality, we use back-propagation in the SMPL-X to adjust pose and shape parameters. The effectiveness of such optimization of parameters was demonstrated in \cite{hu2023gaussianavatar}.

\subsection{Monocular Avatar Training}

\begin{figure}[tb]
  \centering
  \includegraphics[width=0.4\paperwidth]{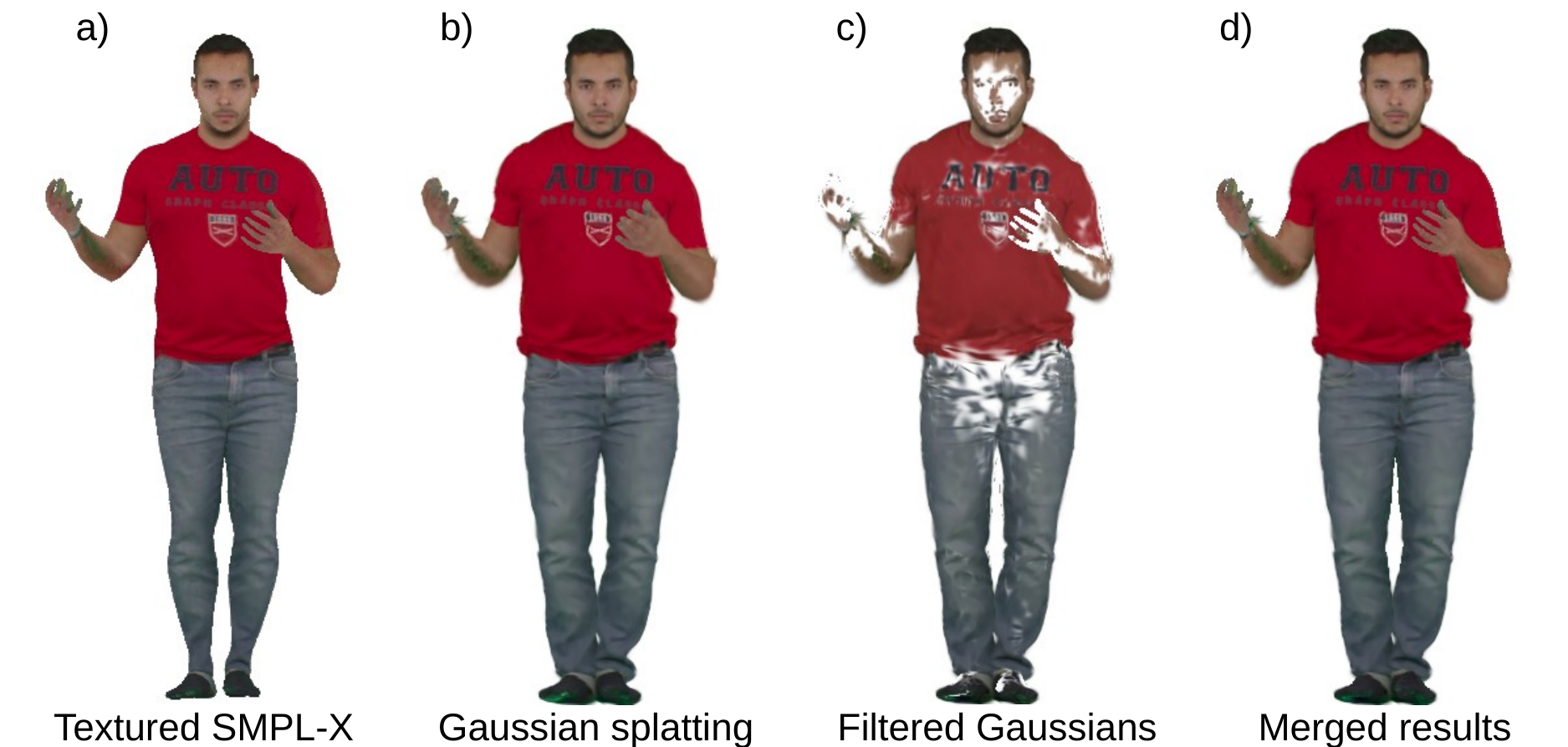}
  \caption{\textbf{Stages of training.} a) SMPL-X with optimizable RGB texture fitted on input video frames. b) 3DGS trained as described in Section \ref{sec:splatting}. c) All unnecessary Gaussians are deleted (Section \ref{sec:merging}) to merge this step with (a) and get (d).}
  \label{fig:stages}
\end{figure}

\textbf{First stage: Gaussian avatar training.} In the first stage (Fig. \ref{fig:scheme} (a)), we train the 3DGS representation of an avatar by  optimizing only local Gaussians transformations $\mu_i^j, r_i^j, s_i^j$ and color $c_i^j$. Opacity $o_i^j$ is fixed to $1$ during this stage as we keep all Gaussians not transparent to efficiently back-propagate image space losses to the SMPL-X parameters. Thus, we force the model to optimize the pose and shape of the underlying mesh rather than deleting Gaussians. We use randomly colored backgrounds in this stage to prevent Gaussians from learning background color. 

To optimize Gaussians we use several image space losses as $L_2$ loss, $L_\textrm{LPIPS}$ perceptual loss \cite{zhang2018perceptual}, $L_\textrm{SSIM}$ structure similarity loss, and $L_\textrm{Sobel}$ loss to get sharper edges. To calculate $L_\textrm{Sobel}$ loss we measure $L_2$ between results of applying the Sobel operator \cite{kanopoulos1988design} to rendered and ground truth images. In other words, we calculate the distance between discrete derivatives of images to account for high-frequency details. We follow \cite{lei2023gart} and apply $L_\textrm{KNN}$, a KNN-based regularization to get smoother results with fewer artifacts. In KNN-regularization we minimize the standard deviation of properties of neighboring Gaussians. The final loss is as follows:
\begin{equation}
    \mathcal{L_{\mathrm{Gaussian}}} = L_2 + \lambda_{LPIPS} L_\textrm{LPIPS} + \lambda_{SSIM} L_\textrm{SSIM} + \lambda_{Sobel} L_\textrm{Sobel} + \lambda_{KNN} L_\textrm{KNN}.
\end{equation}

As a result of this stage, we get a full-body animatable human avatar (Fig. \ref{fig:stages} (b)) with about 25k Gaussians.

\textbf{Second stage: RGB texture training.} In the second stage we render an avatar as rasterized SMPL-X mesh with a texture (Fig. \ref{fig:scheme} (b)). We disable 3DGS and rasterize SMPL-X mesh with trainable texture using Nvdiffrast \cite{laine2020modular}. The use of the differentiable rasterizer lets us back-propagate to the avatar's parameters. We optimize only the texture keeping SMPL-X's parameters frozen during the whole stage. Similar to classic avatar approaches \cite{alldieck2019learning, alldieck2018detailed, alldieck2018video} we utilize three-channeled RGB texture. 

Following \cite{bashirov2024morf}, we utilize TV-regularization ($L_\textrm{TV}$) \cite{chambolle2004algorithm}  to produce smoother results. But we apply $L_\textrm{TV}$ in the texture space instead of the image space as we aim to reduce texture artifacts. The final loss for this stage is as follows:
\begin{equation}
    \mathcal{L_{\mathrm{texture}}} = L_2 + \lambda_{LPIPS} L_\textrm{LPIPS} + \lambda_{SSIM} L_\textrm{SSIM} + \lambda_{TV} L_\textrm{TV}.
\end{equation}

As a result of such training, we get textured mesh (Fig. \ref{fig:stages} (a)) that is fast to render and efficient to store. Although such a representation lacks out-of-mesh details.

\textbf{Third stage:  Filtering out Gaussians.} Textured mesh from the previous stage can replace close-to-surface Gaussians on the avatar (\eg hands and face). Therefore, we can learn which Gaussians to remove (Fig. \ref{fig:stages} (c)) in an unsupervised manner and reduce rendering and storage costs. To achieve this, we merge the differentiable rendering of the textured mesh and the differentiable 3DGS process. 

In Figure \ref{fig:scheme} (c), we render the merged SMPL-X mesh-based and Gaussian avatar (Section \ref{sec:merging}) and train Gaussians opacity $o_i^j$ and color $c_i^j$. We delete all Gaussians with transparency lower than a threshold $(0.1)$. We use two regularizations to encourage optimization to find a trade-off between Gaussians amount and image quality. One reduces the transparency of Gaussians to remove as much of them as possible, while the second preserves Gaussians with a segmentation loss. Using both of them allows us to remove only unnecessary Gaussians.

The transparency regularization pushes opacity $o_i$ of Gaussians down as follows:

\begin{equation}
    L_\textrm{opacity} = \sum_{i=0}^N\sum_{j=0}^{M_i} \lVert o_i^j \rVert_2^2.
\end{equation}

Optimising only this loss would aggressively remove several Gaussians, and for this reason, we add a ``counterweight''. We propose to use silhouette Dice loss ($L_\textrm{dice}$) \cite{milletari2016v}  to encourage the training to preserve out-of-mesh details. As ground truth, we use human silhouettes $S_{\mathrm{GT}}$ that can be predicted by from-the-shelf segmentation models \cite{Gong2019Graphonomy, yang2020renovating}. We summarize alpha map $\mathcal{A}$ and binarized depth map $bin(\mathcal{D})$ to generate silhouette masks for $L_\textrm{dice}$. With these terms, the loss for the third stage is the following:

\begin{equation}
    \mathcal{L_{\mathrm{filtering}}} = \mathcal{L_{\mathrm{Gaussian}}} + \lambda_{opacity}L_\textrm{opacity} + \lambda_{dice}L_\textrm{dice}(S_{\mathrm{GT}}, bin(\mathcal{D}) || \mathcal{A}).
\end{equation}

As a result of such training, we remove only Gaussians that could be replaced with the underlying mesh. Finally, in the inference stage, we utilize preserved Gaussians and trained texture to render an avatar driven by SMPL-X pose parameters.

\subsection{Merging Gaussians with Mesh Representation}
\label{sec:merging}

\begin{figure}[tb]
  \centering
  \includegraphics[width=0.55\paperwidth]{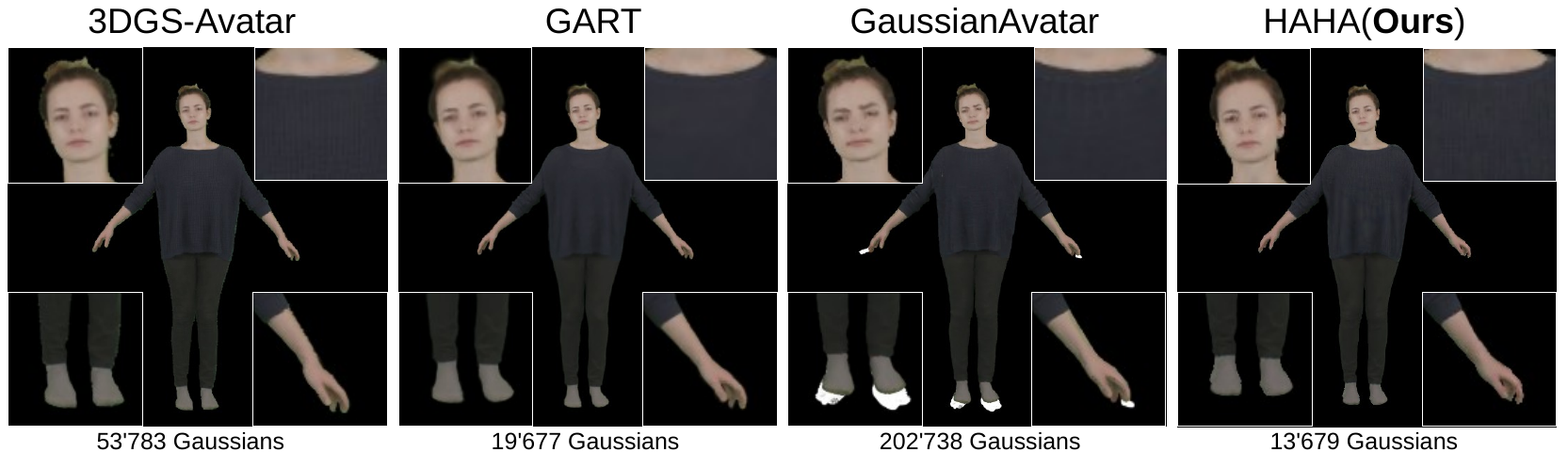}
  \caption{\textbf{Reconstruction for test frames from SnapshotPeople dataset (\textit{female-3-casual}).} Our method demonstrates the same subjective quality of reconstruction as state-of-the-art \cite{qian20233dgs, lei2023gart, hu2023gaussianavatar} while using fewer Gaussians to represent an avatar. For some sequences, GaussianAvatar \cite{hu2023gaussianavatar} tends to include the white background color used in the training while the overall quality of the method is high.}
  \label{fig:reconstruction}
\end{figure}

Here we describe how to simultaneously render 3DGS and textured mesh in a differentiable way. When rendering the textured mesh in Figure \ref{fig:scheme} (b), we calculate its depth map $\mathcal{D}$ as the distance from the camera. We use this depth map as additional input to our modified 3DGS rasterizer $G_{2D}(\mathcal{D}, K, M, \{r', \mu', s', c, o\})$, that also accepts camera intrinsic $K$ and extrinsic $M$ matrices and optimized Gaussians parameters. 

During rasterization, we calculate the distance $D_i$ from the camera to each point of $i$-th Gaussian in the scene. The corresponding value of $D_i$ can be addressed via its screen space coordinates $[x, y]$. Our modification of splatting takes into account the distance $D_i$ in each pixel and compares it to the depth map $\mathcal{D}$ \ie we check if the Gaussians are under the mesh or behind it. We set Gaussian's transparency at each pixel to zero if the distance to Gaussian at this point is more than the depth map value:
\begin{equation}
\label{eq:transparency}
\alpha_i'[x, y] = 
    \left\{ 
    \begin{array}{ll}
        0 & , if D_i[x, y] > \mathcal{D}[x, y]\\
        \alpha_i[x, y] & , else
    \end{array} 
    \right.
,
\end{equation}
where $\alpha_i[x, y]$ initially calculates based on the opacity $o_i^j$ and the Gaussian attenuation (For more details, please refer to the supplementary materials). We also store the final Gaussians transparency map for each pixel to the alpha map $\mathcal{A}$. To do this, we accumulate transparency at each $[x, y]$ pixel during 3DGS rasterization \cite{kerbl20233d}:
\begin{equation}
    \mathcal{A}[x, y] = 1 - \prod_{i=0}^{N[x, y]} (1 - \alpha_i'[x, y]).
\end{equation}
We then use alpha map $\mathcal{A}$ to mix rasterization $\mathcal{M}$ of the textured mesh \mbox{(Fig. \ref{fig:stages} (a))} with Gaussians rasterization $\mathcal{G}$ (Fig. \ref{fig:stages} (c)) to get final avatar (Fig. \ref{fig:stages} (d)). To obtain the final rasterization, we mix them as shown in Figure \ref{fig:scheme}:

\vspace{-0.5em}
\begin{equation}
    \mathcal{I} = \mathcal{G}\mathcal{A} + \mathcal{M}(1-\mathcal{A}).
\end{equation}

The result is coherent because we already set transparency in $\mathcal{A}$ to zero for the Gaussians inside or behind the mesh. This formalization is a fully differentiable pipeline for rendering a mixture of Gaussians and textured mesh. The only issue with such an approach is that 3DGS accumulates color along the ray, taking into account the background color of the scene. We use rasterization $\mathcal{M}$ pixel values as background colors to circumvent color artifacts. So for half-transparent Gaussians we calculate the final color correctly. 

\section{Experiments}
\label{sec:experiments}

In our experiments, we compared HAHA to the state-of-the-art Gaussian methods, namely: GART \cite{lei2023gart}, 3DGS-Avatar \cite{qian20233dgs}, and GaussianAvatar \cite{hu2023gaussianavatar}. All these methods represent the human body as a set of Gaussians. 
We used two open datasets to evaluate our approach: X-Humans \cite{shen2023xavatar} and SnapshotPeople \cite{alldieck2018video}. From both datasets, we used monocular RGB videos as input to our method. In the following section, we show both qualitative and quantitative results.

\begin{figure}[tb]
  \centering
  \includegraphics[width=0.55\paperwidth]{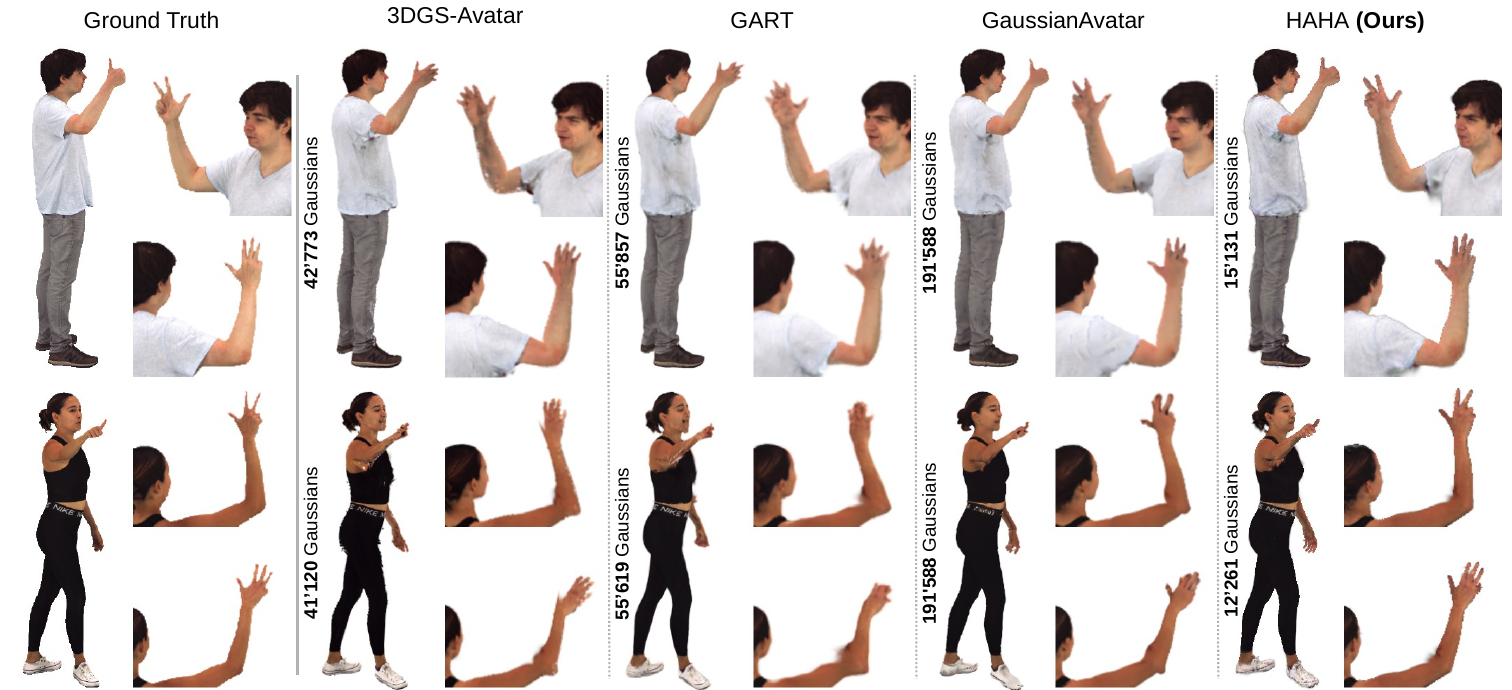}
  \caption{\textbf{Comparison on X-Humans dataset.} We provide results for three different poses and views to demonstrate hands animation. HAHA allows us to animate hands while we use much fewer Gaussians, and it is more robust to the input data while producing fewer artifacts. While GaussianAvatar \cite{hu2023gaussianavatar} also benefits from using SMPL-X to animate hands, HAHA produces more realistic-looking results.}
  \label{fig:xhuman}
\end{figure}

\subsection{Implementation Details}
\label{seq:implementation}
In our experiments, we set loss weight values as follows. We set $\lambda_{LPIPS} = 0.01, \lambda_{SSIM} = 0.1, \lambda_{Sobel} = 1.0, \lambda_{KNN} = 0.01, L_\mathrm{TV}$ to $\lambda_{TV} = 0.01$ for all stages' losses \ie $\mathcal{L_{\mathrm{Gaussian}}}$, $\mathcal{L_{\mathrm{texture}}}$ and $\mathcal{L_{\mathrm{filtering}}}$. We get the best trade-off in quality and number of Gaussians for learnable removing with the following regularization weights: $\lambda_{opacity}=0.001, \lambda_{dice}=0.1$.
We trained all avatars using batch size equal to 4 using Adam \cite{Kingma2014AdamAM} optimizer. In the first stage of training, we optimize Gaussian parameters for 3000 iterations. Then we optimize texture in the second stage for 2500 iterations. In the last stage, we fine-tuned Gaussians' color and opacity for 5000 iterations. 
We set other hyperparameters (such as learning rates) similar to GART \cite{lei2023gart}. For more details, please refer to supplementary materials. 

\begin{table}[tb]
\caption{Quantitative metrics for X-Humans \cite{shen2023xavatar} dataset. The dataset lets one evaluate metrics values for novel poses.}
\label{tab:xhumans}
\fontsize{8}{10}\selectfont
\centering
\begin{tabular}{l|llll|llll}
            & \multicolumn{4}{c|}{00016 (male)} & \multicolumn{4}{c}{00019 (female)} \\ \hline
            & Gaussians$\downarrow$    & PSNR$\uparrow$     & SSIM$\uparrow$    & LPIPS$\downarrow$   & Gaussians$\downarrow$   & PSNR$\uparrow$     & SSIM$\uparrow$     & LPIPS$\downarrow$      \\ \hline
3DGS-Avatar \cite{qian20233dgs}  &  \cellcolor{yellow!25}42.77k  &  25.44  &  0.9315  &  \cellcolor{green!25}0.0409  &  \cellcolor{yellow!25}41.12k  &  27.63  &  0.9539  &  \cellcolor{green!25}0.0471  \\  
GART  \cite{lei2023gart}  &  55.85k  &  \cellcolor{green!25}25.71  &  0.9295  &  0.0598  &  55.61k  &  \cellcolor{yellow!25}27.78  &  0.9512  &  0.0668  \\
GaussianAvatar \cite{hu2023gaussianavatar}  &  191.58k  &  \cellcolor{yellow!25}25.58  &  \cellcolor{yellow!25}0.9328  &  0.0518  &  191.58k  &  27.54  &  \cellcolor{yellow!25}0.9574  &  0.0647 \\ \hline
HAHA(\textbf{Ours})  &  \cellcolor{green!25}15.13k  &  25.49  &  \cellcolor{green!25}0.9339  &  \cellcolor{yellow!25}0.0507  &  \cellcolor{green!25}12.26k  &  \cellcolor{green!25}28.49  & 
 \cellcolor{green!25}0.9593  &   \cellcolor{yellow!25}0.0501         
\end{tabular} 

\vspace*{0.5\baselineskip}

\begin{tabular}{l|llll|llll}
            & \multicolumn{4}{c|}{00018 (male)} & \multicolumn{4}{c}{00027 (female)} \\ \hline
            & Gaussians$\downarrow$    & PSNR$\uparrow$     & SSIM$\uparrow$    & LPIPS$\downarrow$   & Gaussians$\downarrow$   & PSNR$\uparrow$     & SSIM$\uparrow$     & LPIPS$\downarrow$      \\ \hline
3DGS-Avatar \cite{qian20233dgs}  &  \cellcolor{yellow!25}26.78k  &  28.71  &  0.9521  &  \cellcolor{yellow!25}0.0580  &  36.82k  &  \cellcolor{yellow!25}26.84  &  0.9477  &  \cellcolor{green!25}0.0445  \\
GART  \cite{lei2023gart}  & 50.47k  &  \cellcolor{yellow!25}30.98  &  \cellcolor{yellow!25}0.9595  &  0.0683  &  \cellcolor{yellow!25}47.18k  &  26.56  &  0.9449  &  0.0595  \\
GaussianAvatar \cite{hu2023gaussianavatar}  & 191.58k   & 29.92   & 0.9588   & 0.0744  & 191.58k & 25.69  &  \cellcolor{yellow!25}0.9481 & 0.0543   \\ \hline
HAHA(\textbf{Ours})  &  \cellcolor{green!25}18.57k  &  \cellcolor{green!25}31.10  &  \cellcolor{green!25}0.9630  &  \cellcolor{green!25}0.0579  &  \cellcolor{green!25}15.50k  &  \cellcolor{green!25}27.26  &  \cellcolor{green!25}0.9513  &  \cellcolor{yellow!25}0.0473         
\end{tabular}
\end{table}


\begin{table}[tb]
\caption{Quantitative metrics for SnapshotPeople \cite{alldieck2018video} dataset. Our method gets metrics on par with state-of-the-art approaches while using much fewer Gaussians.}
\label{tab:snapshot}
\fontsize{8}{10}\selectfont
\centering
\begin{tabular}{l|llll|llll}
            & \multicolumn{4}{c|}{female-3-casual} & \multicolumn{4}{c}{male-3-casual} \\ \hline
            & Gaussians$\downarrow$    & PSNR$\uparrow$     & SSIM$\uparrow$      & LPIPS$\downarrow$    & Gaussians$\downarrow$   & PSNR$\uparrow$     & SSIM$\uparrow$     & LPIPS$\downarrow$     \\ \hline
3DGS-Avatar \cite{qian20233dgs}  &  53.78k  &  30.57  &  0.9581  &  \cellcolor{green!25}0.0208  &  37.22k  &  \cellcolor{yellow!25}34.28  &  \cellcolor{yellow!25}0.9724  & \cellcolor{green!25}0.0149  \\ 
GART \cite{lei2023gart}  &  \cellcolor{yellow!25}19.67k  &  \cellcolor{green!25}32.73  &  \cellcolor{yellow!25}0.9672  &  0.0459  &  \cellcolor{yellow!25}21.88k  &  \cellcolor{green!25}35.93  &  \cellcolor{green!25}0.9767   &  0.0294 \\
GaussianAvatar \cite{hu2023gaussianavatar}  & 202.73k  &  25.94  &  \cellcolor{green!25}0.9673  &  0.0434 &  202.73k  &  33.59  &  0.9697  &  \cellcolor{yellow!25}0.0243  \\ \hline
HAHA(\textbf{Ours})  &  \cellcolor{green!25}13.67k  &  \cellcolor{yellow!25}32.53  &  0.9633  & \cellcolor{yellow!25}0.0403  &  \cellcolor{green!25}13.60k  &  31.46  &  0.9619  &  0.0277   
\end{tabular} 
\end{table}

\subsection{X-Humans} 
We report the following metrics: PSNR, SSIM, and LPIPS \cite{zhang2018unreasonable} (Table \ref{tab:xhumans}). PSNR and SSIM measure the fidelity of the signal and structural similarity, respectively, while LPIPS correlates with human perception of the image using neural network features to compare with ground truth. We evaluated metrics on the renderings with a black background as in 3DGS-Avatar \cite{qian20233dgs} experiments to set the background value to zero. During inference, we used \textit{test time pose optimization} following GART \cite{lei2023gart} to reduce the impact of SMPL-X fitting inaccuracies.

X-Humans \cite{shen2023xavatar} dataset provides a sequence of frames with rendered 3D scans of a person doing complex movements. The movements are diverse for both training and testing videos, therefore it is a challenging task to train on such a dataset. In Table \ref{tab:xhumans} we compare our method with previous state-of-the-art methods: GART \cite{lei2023gart}, 3DGS-Avatar \cite{qian20233dgs}, and GaussianAvatar \cite{hu2023gaussianavatar}. The last one, similar to us, uses SMPL-X and can control fingers animation so we can compare the animation of hands. We provide metrics for both male and female avatars.

HAHA is more robust and gets better metrics for these complex training and testing sequences. Besides, our method requires fewer Gaussians. We also provide qualitative results in Figure \ref{fig:xhuman} demonstrating overall quality and how our approach handles hands animation. Additional visualizations for more people from the dataset can be found in the supplementary materials.

\begin{figure}[tb]
  \centering
  \includegraphics[width=0.55\paperwidth]{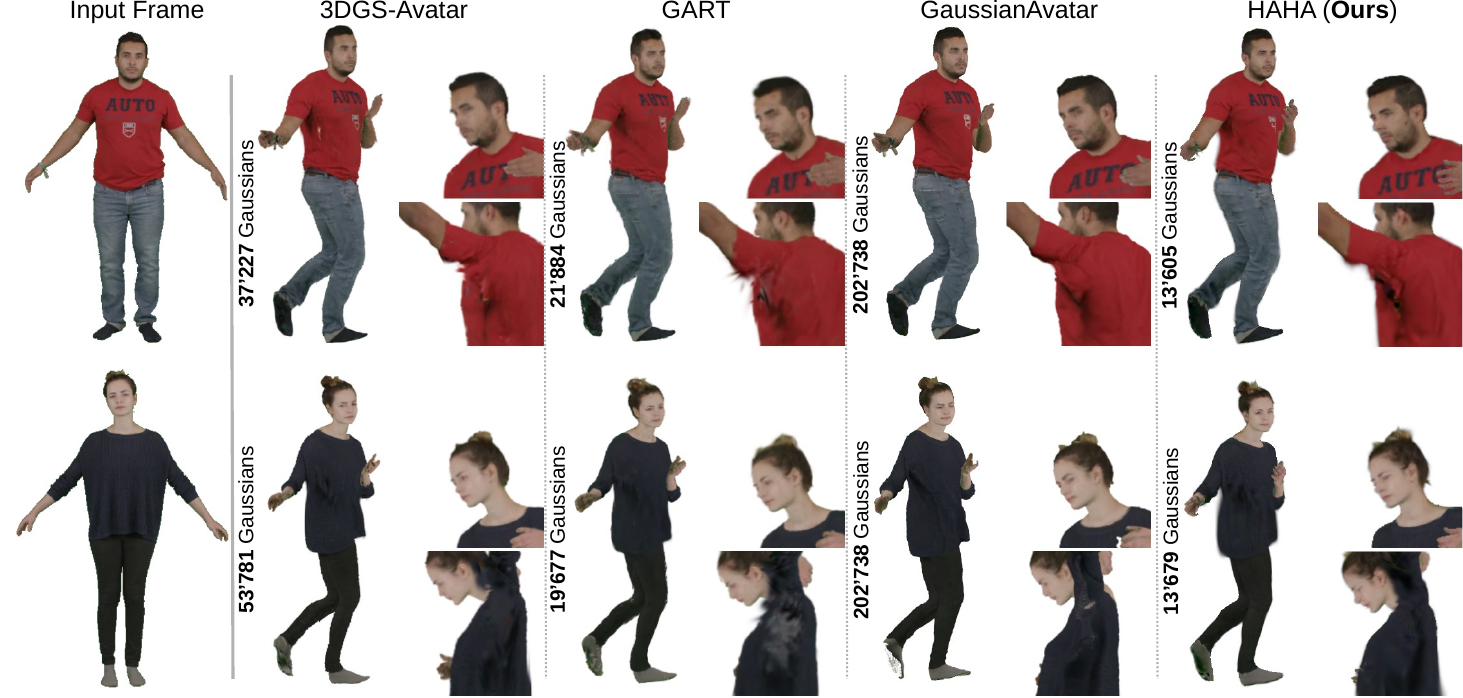}
  \caption{\textbf{Novel poses for SnapshotPeople dataset.} HAHA reduces the number of artifacts for novel views and body regions unseen during training. At the same time, we use fewer Gaussians for rendering.}
  \label{fig:results}
\end{figure}

\subsection{PeopleSnapshot} 
Following the previous literature, we also provide quantitative metrics for the SnapshotPeople \cite{alldieck2018video} dataset (Table~\ref{tab:snapshot}). However, SnapshotPeople does not allow assess quality for novel views and poses since train and test sequences are very similar-looking.

In all experiments for SnapshotPeople we used SMPL provided by AnimNerf~\cite{chen2021animatable}. As our method requires a parametric model to have articulated fingers, we converted the provided SMPL to SMPL-X using a converter from the SMPL official repository. Then we fine-tuned the resulting SMPL-X hand's pose and shape using SMPLify-X\cite{pavlakos2019expressive} to match ground truth frames. Similar to X-Humans experiments, we evaluate metrics with a black background and use \textit{test time pose optimization} during inference.

SnapshotPeople evaluation methodology is challenging for our method because we strongly rely on the underlying mesh geometry. Therefore, in cases when train and test views and poses are similar, we could face metrics value reduction on the opposite to the methods where rendering does not strongly depend on the mesh surface. Nevertheless, we demonstrate metrics on par with state-of-the-art approaches for this dataset while using almost two times fewer Gaussians (Table \ref{tab:snapshot}). In Figure \ref{fig:reconstruction} we provide a qualitative comparison of avatar reconstruction for test frames from the PeopleSnapshot dataset. Our method demonstrates on par quality of the avatar using fewer Gaussians. Additionally, in some regions that are difficult to represent with Gausssians based on limited input data, our method reduces the number of artifacts (Fig. \ref{fig:results}).

The qualitative improvement to the state-of-the-art is noticeable for novel poses and viewpoints (Fig. \ref{fig:results}). To demonstrate how our and competitors' methods handle novel poses, we provide a comparison with reposed results. The use of textured mesh prior not only allows us to reduce the number of Gaussians but also reduces artifacts, especially in areas not sufficiently represented in the training frames. Additional visualizations and metrics for more people from the dataset can be found in the supplementary materials.

\subsection{Ablation Study.}

\begin{table}[tb]
\caption{Ablation study of losses and regularizations on \textit{female-4-casual} from SnapshotPeople.}
\label{tab:ablation_acc}
\centering
\begin{tabular}{l|llll}
                      & PSNR $\uparrow$ & SSIM $\uparrow$ & LPIPS $\downarrow$   & Gaussians $\downarrow$   \\ \hline
No Sobel loss                     & 30.74  & 0.9564 & 0.0331 & \cellcolor{yellow!25}6.89k    \\ 
No opacity regularization         & \cellcolor{yellow!25}30.95  & \cellcolor{yellow!25}0.9582 & \cellcolor{yellow!25}0.0289 & 22.56k    \\
No segmentation regularization    & 28.80  & 0.9450 & 0.0364 & \cellcolor{green!25}2.00k    \\
\hline
Full pipeline         & \cellcolor{green!25}31.15   & \cellcolor{green!25}0.9589  & \cellcolor{green!25}0.0283  & 11.96k               
\end{tabular} 
\end{table}

\begin{figure}[tb]
  \centering
  \includegraphics[width=0.45\paperwidth]{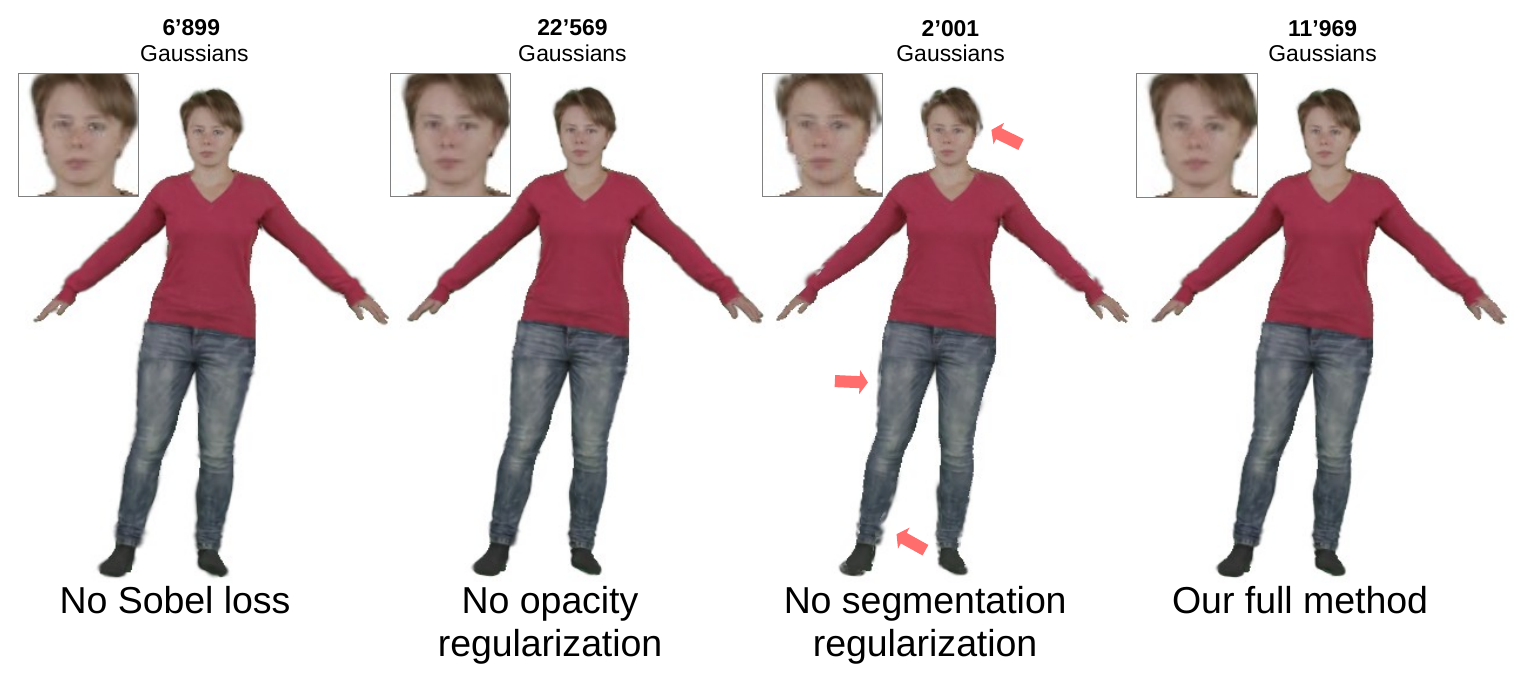}
  \caption{\textbf{Losses ablation study.} HAHA gets the best trade-off between image quality and the number of Gaussians with the full set of proposed losses. Also removing Sobel loss could help to additionally reduce the amount of Gaussians but it leads to blurred edges.}
  \label{fig:ablation}
\end{figure}

First, we ablate our loss choices: Sobel loss and the two proposed opacity regularizations. In Table \ref{tab:ablation_acc}, we provide quantitative metrics to evaluate the impact of each design choice. Figure \ref{fig:ablation} shows avatars corresponding to each table's row. According to our experiments, Sobel loss acts as an additional regularizer that prevents the deletion of Gaussians as it restricts the preserving of high-frequency details. So it could be switched off if one needs even fewer Gaussians but this will affect the sharpness of the edges and values of the quantitative metrics.

Both opacity and segmentation regularizations are essential to control the Gaussians amount. Without opacity regularization the method tends to inefficient Gaussians removal. The removal of segmentation regularization results in deleting too many Gaussians causing severe artifacts. We conclude that one should use both these regularizations simultaneously to get the best result.

We also evaluated the quality of avatars after each stage of training. We provided quantitative metrics in Table \ref{tab:merging_acc} while qualitative comparison can be found in Figure \ref{fig:stages} (a, b, d). In this experiment, we also ablated the effectiveness of unnecessary Gaussians removal. \textit{Naive merging} is a baseline method that merges Gaussians with a textured mesh without filtering them out. Using more Gaussians leads to higher PSNR and SSIM, while LPIPS decreases. As LPIPS is known for better correlation with human perception, we claim that removing part of the Gaussians leads to better quality.

\begin{table}[tb]
\caption{Metrics for each stage of avatar training pipeline. Metrics evaluated for \textit{00019} subject from X-Humans dataset.}
\label{tab:merging_acc}
\fontsize{8}{10}\selectfont
\centering
\begin{tabular}{l|cccc|lll}
                      & Gaussians $\downarrow$ & Storage $\downarrow$ & FPS$\uparrow$ & Train $\downarrow$ & PSNR $\uparrow$ & SSIM $\uparrow$ & LPIPS $\downarrow$     \\
                      &    & (Mb) & (Inference) & (min:sec) &  &  &  \\\hline
Textured mesh         & ------    & 0.196 &  $463.58 \pm 17.46$ & 5:24 & 27.26           & 0.9554          & 0.0539                      \\ \hline
Full-Gaussian         & \cellcolor{yellow!25}37.25k    & \cellcolor{yellow!25}2.086   & \cellcolor{yellow!25}$240.09 \pm 7.65$ & 7:58 & \cellcolor{green!25}28.66  & \cellcolor{green!25}0.9601 & 0.0572                    \\ 
Naive merging         & 37.25k    & 2.282  & $238.17 \pm 7.81$ & ------ & \cellcolor{yellow!25}28.51           & \cellcolor{yellow!25}0.9599          & \cellcolor{yellow!25}0.0510                       \\
\textbf{Finetuning} & \cellcolor{green!25}12.26k & \cellcolor{green!25}0.883 & \cellcolor{green!25}$247.87 \pm 4.62$ & 12:19 & 28.49           & 0.9593          & \cellcolor{green!25}0.0501 
\end{tabular}
\end{table}

In Table \ref{tab:merging_acc} we demonstrate how the reduction in the number of Gaussians affects the storage space consumption and rendering speed. Using fewer Gaussians with 256$\times$256 RGB texture lets us use more than 2.3x less memory to store an avatar. Such storage space reduction could be useful for industrial applications when it is necessary to store avatars for millions of users. At the same time, we demonstrate total convergence speed (25 min 41 sec) on par with other methods: GaussianAvatar (28 min 37 sec), 3DGS-Avatar (26 min 03 sec).


\section{Discussion and Conclusion.}
We have presented a new method for modeling human avatars using joint representation with RGB textured mesh and Gaussian splatting. We use a textured SMPL-X parametric model to portray the avatar's areas near the human body surface while using Gaussians to render out-of-mesh details. Our methods allow us to significantly reduce the number of Guassians and memory required to store avatars. Using textured SMPL-X for body parts representation allows us to animate small details such as fingers. We demonstrated the efficiency of our approach both quantitatively and qualitatively on the open datasets. HAHA outperforms the previous state-of-the-art on challenging X-Humans dataset.

Our method's limitation is the difficulty of getting an accurate SMPL-X mesh for an input video. As we strongly depend on how accurate mesh projection matches the input frames. The task of getting SMPL-X parameters from a monocular video is long-standing but still has room for improvement.

\clearpage  

%
%
\bibliographystyle{splncs04}
\bibliography{main}


\clearpage
\setcounter{section}{0}
\renewcommand{\thesection}{\Alph{section}}

 
\let\titleold\title
\renewcommand{\title}[1]{\titleold{#1}\newcommand{\thetitle}{#1}}
\def\maketitlesupplementary
   {
   \onecolumn
   \begin{center}
   \Large
    \textbf{\thetitle}\\
    \small
    \vspace{0.5em}------------------------------ \\
    \vspace{1.0em}
   \end{center}
   }
 

\title{Supplementary materials for "HAHA: Highly Articulated Gaussian Human Avatars with Textured Mesh Prior"}
\maketitlesupplementary

\section{Further Implementation Details and Algorithms}
\subsection{Hyper-parameter values.} Here, we provide more implementation details in addition to Section \ref{seq:implementation}. In our implementation, we use the following set of \textit{learning rates}. For $\mathcal{L_{\mathrm{Gaussian}}}$ we use: $lr_{color} = 0.005, lr_{scaling} = 0.005, lr_{rotation} = 0.005$ that regulate speed of updating $c_i^j, s_i^j, r_i^j$ respectively. For $\mu_i^j$ we apply \textit{Exponential annealing} of learning rate from $lr_{xyz} = 0.00016$ to $lr'_{xyz}=0.0000016$ during all 3000 training steps for the first stage. We update SMPL-X body and shape parameters with $lr_{pose} = 0.0002$. For $\mathcal{L_{\mathrm{texture}}}$ we use $lr_{texture} = 0.01$ to update texture colors. For $\mathcal{L_{\mathrm{filtering}}}$ we use $lr_{color} = 0.005, lr_{opacity} = 0.05$ for $c_i^j$ and $o_i^j$ respectively.

\subsection{Polygon transformation parameters}
\begin{figure}[tb]
  \centering
  \includegraphics[width=0.35\paperwidth]{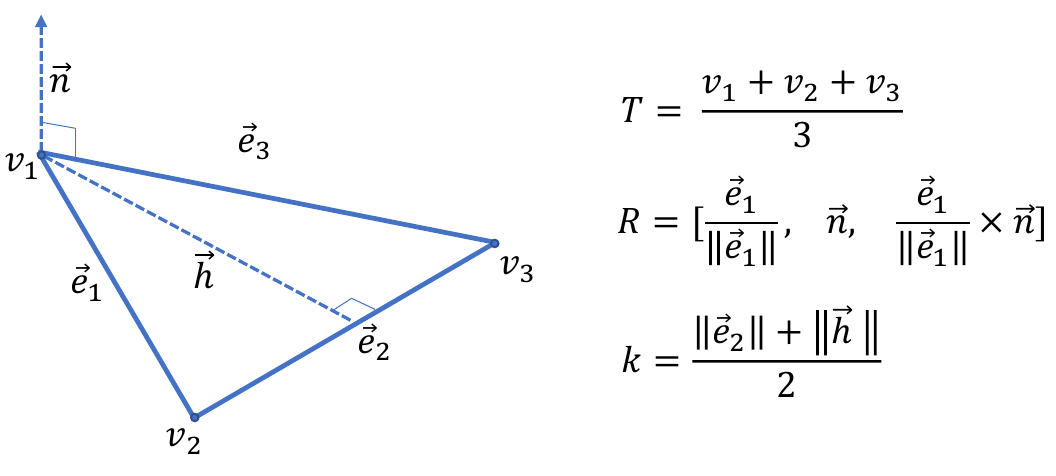}
  \caption{\textbf{Polygon transformation calculation.} We calculate polygon translation $T$, rotation $R$ and scale $k$ as shown in the figure. $T$ is the mean position of the vertices, $R$ is rotation matrix formed based on polygon vectors, $k$ is a scale coefficient obtained similar to area calculation.}
  \label{fig:triangle}
\end{figure}

This section describes how we calculate $\{T_i, R_i, k_i\}$ polygon transformation parameters. We assume that all polygons are triangles and calculate transformation parameters following \cite{qian2023gaussianavatars}. We calculate position $T_i$ as the average coordinate of the polygon's vertices \ie center of the polygon. To calculate scale $k_i$, we calculate the mean between the lengths of one of the edges and the triangle height. This way, $k_i$ correlates with the triangle area. To find quaternion rotation $R_i$, we first find a rotation matrix as a concatenation of the direction vector of one edge, normal vector, and their cross product, then convert such a matrix to quaternion representation. Figure \ref{fig:triangle} shows the described polygon transformations calculation.

\begin{figure}[tb]
  \centering
  \includegraphics[width=0.55\paperwidth]{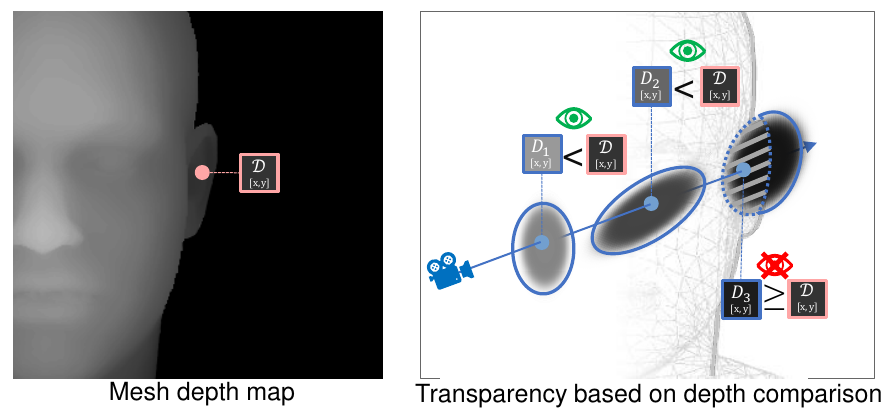}
  \caption{\textbf{Gaussian transparency calculation.} We calculate the Gaussian transparency at each point, comparing the corresponding Gaussians' depths along the ray to the mesh depth map value at a given pixel $[x, y]$.} 
  \label{fig:depth}
\end{figure}

\subsection{Depth-conditioned transparency}
We provide a more detailed explanation of transparency calculation for equation \ref{eq:transparency}. In the following, we use the notation introduced in Section \ref{sec:merging}. In Figure \ref{fig:depth}, we provide a scheme of Gaussian transparency calculation. First, we check the mesh depth map $\mathcal{D}$ value at each pixel $[x, y]$ to calculate the transparency of Gaussians at that point (Fig. \ref{fig:depth} (left)). Then, during 3DGS rasterization, we compare for each $i$-th Gaussian along the ray its depth $D[x, y]$ with respect to $\mathcal{D}[x, y]$ (Fig. \ref{fig:depth} (middle)). We set Gaussian transparency $\alpha_i[x, y]$ for $i$-th Gaussian at point $[x, y]$ to zero if the depth of the Gaussian at this point is more than the stored mesh depth ($D_i[x, y] < \mathcal{D}[x, y]$). As a result, during rasterization, we not showing parts of Gaussians with $\alpha_i[x, y] == 0$.

\section{Additional Comparisons}

\begin{figure}[tb]
  \centering
  \includegraphics[width=0.55\paperwidth]{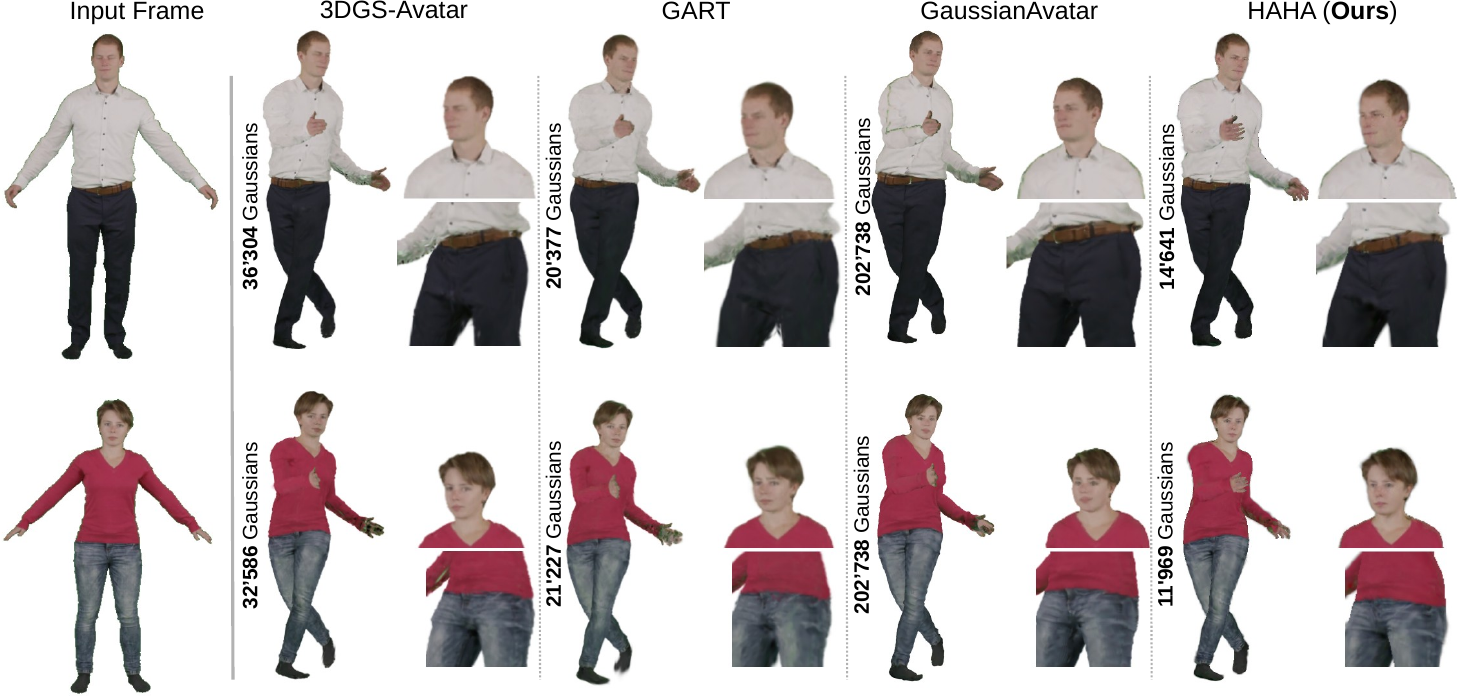}
  \caption{\textbf{Additional visualizations for novel poses for SnapshotPeople dataset}. HAHA generates less blurred results with fewer artifacts for high-frequency arias such as faces.} 
  \label{fig:snapshot_other}
\end{figure}

In this section we provide additional qualitative results for SnapshotPeople \cite{alldieck2018video} and X-Humans \cite{shen2023xavatar} datasets. We provide a comparison for more subjects from these datasets and additional quantitative metrics for SnapshotPeople. 



\textbf{Qualitative results for SnapshotPeople.} Here we provide quantitative (Table \ref{tab:snapshot}) and qualitative (Fig. \ref{fig:snapshot_other}) results for more people from SnapshotPeople dataset. In Figure \ref{fig:snapshot_other}, we provide qualitative visualization in novel poses for two more people commonly used for comparison. All previous state-of-the-art methods use these four subjects (Fig. \ref{fig:results}, \ref{fig:snapshot_other}) for comparison because they have SMPL fits provided by AnimNerf \cite{chen2021animatable}. HAHA generalizes to the novel poses and views producing sharp and consistent details (\eg belt or face for the top row, edge of sweater for bottom row).

\begin{figure}[tb]
  \centering
  \includegraphics[width=0.55\paperwidth]{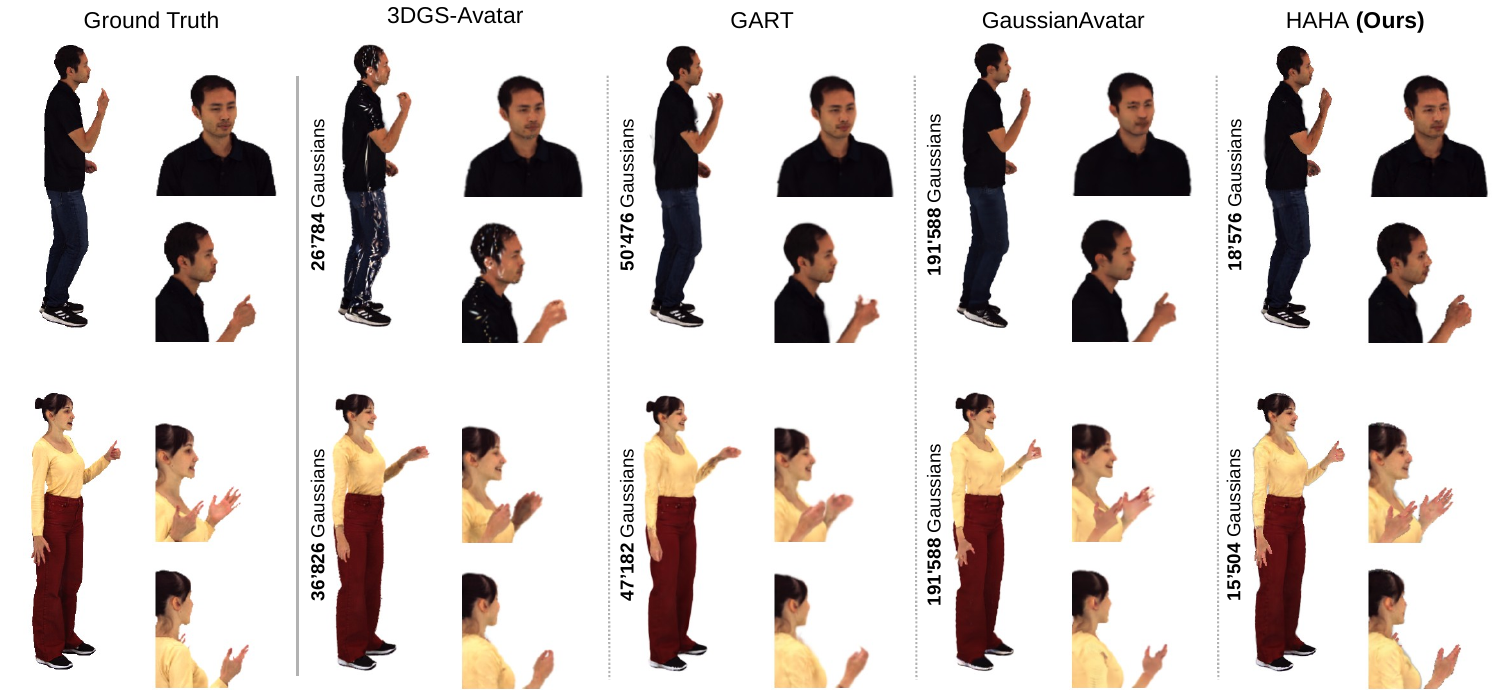}
  \caption{\textbf{Additional visualizations for X-Humans dataset.} HAHA produces less blurred faces than previous state-of-the-art and better preserves clothes details like lines on the sneakers.}
  \label{fig:xhuman_other}
\end{figure}

\textbf{Qualitative results for X-Humans.} We provide additional visualizations for X-Humans \cite{shen2023xavatar} dataset in Figure \ref{fig:xhuman_other}. These visualizations demonstrate renders for two more people (00018 and 00027) from Table \ref{tab:xhumans}. The first two people are visualized in Figure \ref{fig:xhuman} in the main paper. Presented here qualitative results confirm the conclusion provided in the main paper. 

\section{Gaussian Density Analysis}
We carried out density analysis to demonstrate the efficiency of learnable Gaussians removal. We define density based on the average distance to the six nearest neighbors for each Gaussian. After calculating such distance for each Gaussian, we normalize and invert values over the avatar to obtain the mapped density.

\begin{figure}[tb]
  \centering
  \includegraphics[width=0.55\paperwidth]{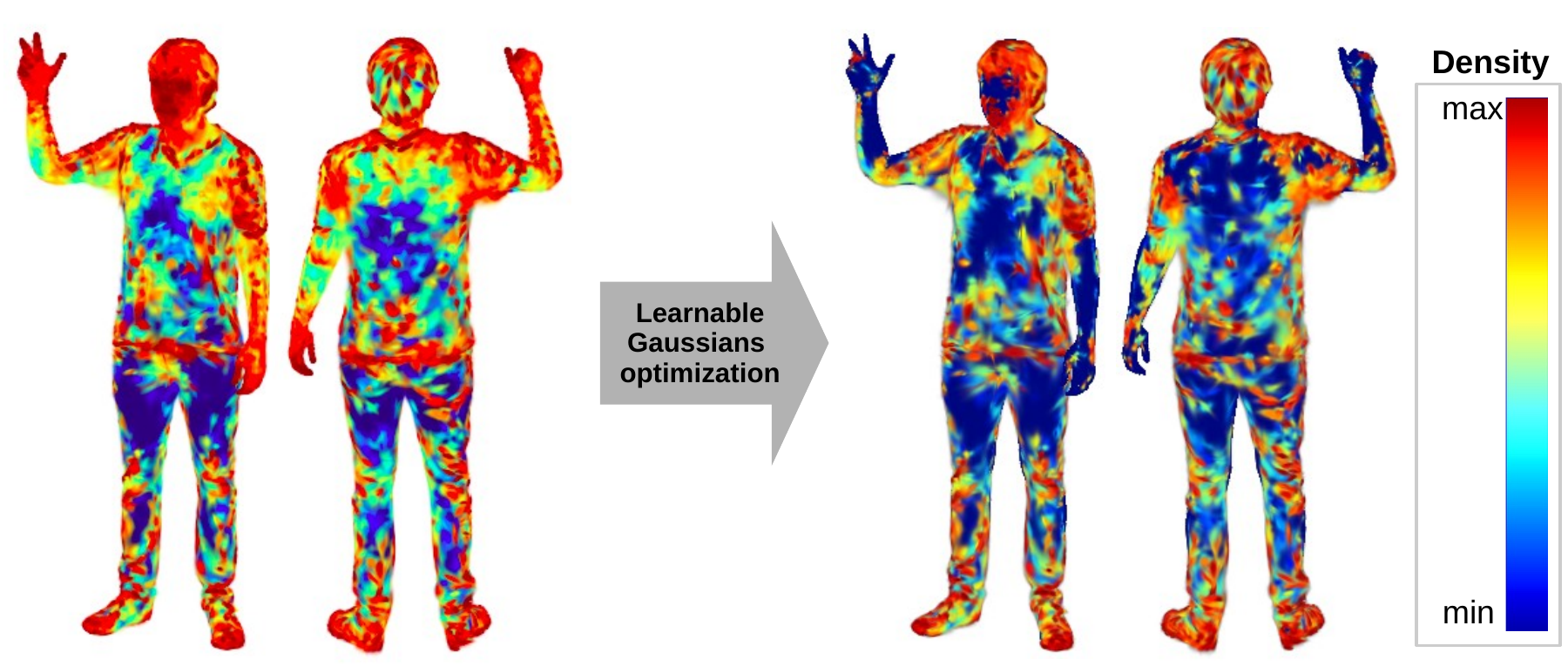}
  \caption{\textbf{Gaussians density visualization}. Visualized Gaussians' density as a pseudo-colored heatmap demonstrates that HAHA removes the most dense areas (in red) and makes Gaussians more sparse (in blue). We provide both front and back views using different frames from the test sequence.}
  \label{fig:density}
\end{figure}

In Figure \ref{fig:density}, we provide a heatmap visualization of Gaussians' density. On the left, we demonstrate the densities of Gaussians for front and back views before learnable removal. This additional pseudo-colored visualization demonstrates that HAHA successfully replaces the most dense regions with the textured mesh. In particular, highly articulated areas such as fingers and faces can be replaced with mesh representation using HAHA.

We also noticed that HAHA keeps dense regions in cases where it is necessary to represent complex out-of-mesh details (\eg hairs and feet). While in cases where rendering result is not so sensitive to Gaussians' density, HAHA makes representation sparser (\eg close to body areas of a T-shirt). We conclude that HAHA is well suited for human avatar tasks and can efficiently reduce the number of Gaussians without causing significant quality reduction.

\begin{table}[tb]
\caption{Quantitative metrics for two more people from SnapshotPeople \cite{alldieck2018video} dataset.}
\label{tab:snapshot}
\fontsize{8}{10}\selectfont
\centering

\begin{tabular}{l|llll|llll}
            & \multicolumn{4}{c|}{male-4-casual} & \multicolumn{4}{c}{female-4-casual} \\ \hline
            & Gaussians$\downarrow$    & PSNR$\uparrow$      & SSIM$\uparrow$      & LPIPS$\downarrow$  & Gaussians$\downarrow$  & PSNR $\uparrow$     & SSIM$\uparrow$      & LPIPS$\downarrow$    \\ \hline
3DGS-Avatar \cite{qian20233dgs}  &  35.96k  &  30.22  &  0.9653  &  \cellcolor{green!25}0.0230  &  33.44k  &  \cellcolor{yellow!25}33.16  &  \cellcolor{yellow!25}0.9678  &  \cellcolor{green!25}0.0157  \\ 
GART \cite{lei2023gart}  &  \cellcolor{yellow!25}20.37k  &  \cellcolor{green!25}33.17  &  \cellcolor{green!25}0.9719  &  0.0376  &  \cellcolor{yellow!25}21.22k  &  \cellcolor{green!25}34.42  &  \cellcolor{green!25}0.9706  &  0.0328  \\
GaussianAvatar \cite{hu2023gaussianavatar}  &  202.73k  &  \cellcolor{yellow!25}30.83  &  \cellcolor{yellow!25}0.9676  &  \cellcolor{yellow!25}0.0302  &  202.73k  &  32.98  &  0.9676  &  \cellcolor{yellow!25}0.0239  \\ \hline
HAHA(\textbf{Ours})  &  \cellcolor{green!25}14.64k  &  27.08  &  0.9505  & 0.0432  &  \cellcolor{green!25}11.96k  &  31.15  &  0.9589  &  0.0283   
\end{tabular}

\end{table}


\end{document}